\pdfoutput=1

\documentclass[11pt]{article}
\usepackage{arabtex}
\usepackage[]{EMNLP2023}
\usepackage{lipsum}
\usepackage{times}
\usepackage{latexsym}

\usepackage[T1]{fontenc}

\usepackage[utf8]{inputenc}
\usepackage{xstring}
\usepackage{cleveref}
 \usepackage{arabtex}
\usepackage{utf8}
\usepackage{color}
\usepackage{arydshln}
\usepackage{adjustbox}
\usepackage{float}
\restylefloat{table}
\usepackage{array,booktabs,makecell}
\usepackage{multirow}
\usepackage{graphicx}
\usepackage{color, colortbl}
\usepackage{xstring}
\usepackage{arydshln}
\usepackage{microtype}
\usepackage{color,soul}

\usepackage{multirow}
\usepackage{graphicx}

\usepackage{listings}
\usepackage{color}

\definecolor{dkgreen}{rgb}{0,0.6,0}
\definecolor{gray}{rgb}{0.5,0.5,0.5}
\definecolor{mauve}{rgb}{0.58,0,0.82}

\newcommand\blfootnote[1]{%
  \begingroup
  \renewcommand\thefootnote{}\footnote{#1}%
  \addtocounter{footnote}{-1}%
  \endgroup
}

\usepackage{inconsolata}
\usepackage{arydshln}
\newcolumntype{H}{>{\setbox0=\hbox\bgroup}c<{\egroup}@{}}
\usepackage{cancel}
\usepackage{ulem,xpatch}

\newcommand{\octopus}{\textsc{Octopus}}
\usepackage{hyperref}
\def\r{\color{red}}
\definecolor{blush}{rgb}{0.87, 0.36, 0.51}
\def\b{\color{blush}}
\definecolor{dgreen}{rgb}{0.0, 0.5, 0.0}
\def\g{\color{dgreen}}
\def\n{\color{black}}
\def\o{\color{orange}}
\def\u{\color{blue}}

\title{\includegraphics[scale=0.2]{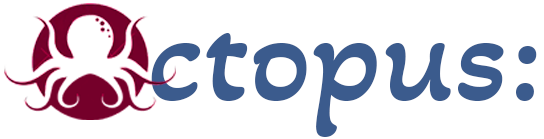} \\ A Multitask Model and Toolkit for Arabic Natural Language Generation}

\author{\normalsize AbdelRahim Elmadany$^{\xi,\star}$ ~ El Moatez Billah Nagoudi$^{\xi,\star}$ ~ Muhammad Abdul-Mageed$^{\xi,\lambda,\star}$ \\
\normalsize $^{\xi}$ Deep Learning \& Natural Language Processing Group,
  The University of British Columbia\\\normalsize  $^{\lambda}$Department of Natural Language Processing \& Department of Machine Learning, MBZUAI\\ %
  \texttt{\normalsize \{a.elmadany,moatez.nagoudi,muhammad.mageed\}@ubc.ca}}
  
\setcode{utf8}
\setarab 
\begin{document}
\maketitle

\begin{abstract}

Understanding Arabic text and generating human-like responses is a challenging endeavor. While many researchers have proposed models and solutions for individual problems, there is an acute shortage of a comprehensive Arabic natural language generation toolkit that is capable of handling a wide range of tasks. In this work, we present a novel Arabic text-to-text Transformer model, namely AraT5\textsubscript{v2}. Our new model is methodically trained on extensive and diverse data, utilizing an extended sequence length of $2,048$ tokens. We explore various pretraining strategies including unsupervised, supervised, and joint pertaining, under both single and multitask settings. Our models outperform competitive baselines with large margins. We take our work one step further by developing and publicly releasing~\octopus, a Python-based package and command-line toolkit tailored for \textit{eight} Arabic generation tasks all exploiting a \textit{single} model. We release the models and the toolkit on our public repository.\footnote{\href{https://github.com/UBC-NLP/octopus}{https://github.com/UBC-NLP/octopus}}

\end{abstract}
\section{Introduction}
 ~\blfootnote{ $^{\star}${Equal contributions}}

Natural Language Generation (NLG) is a fundamental component of natural language processing that aims to generate human-like, coherent, contextually fitting, and linguistically precise text from structured data or various other input formats. NLG systems find applications in various aspects of daily life, including education, healthcare, business, and more. The recent emergence of generative models has significantly impacted the field of NLG. 
While important progress has been made in NLG research, the majority of existing tools, systems, and models are primarily focused on English~\cite{jhaveri2019clstk, khan2021generate, lauriola2022introduction}, leaving behind many languages, including Arabic. 

Although it is one of the most widely spoken languages in the world, and one with a rich linguistic structure and diverse dialects, Arabic remains underrepresented in NLG. One reason is the complex morphology and syntax of Arabic. Hence, the primary focus of our research here is to develop an advanced tool capable of performing several key Arabic NLG tasks. For example, we target tasks such as \textit{text summarization}, \textit{question answering}, \textit{question generation}, \textit{news headline generation}, and \textit{paraphrasing}. These are tasks that necessitate a deep understanding of semantics, syntax, and pragmatics of Arabic. We also focus on tasks that require an understanding of both the syntax and morphology such as \textit{diacritization}, \textit{transliteration}, and \textit{grammatical error correction}. Our main contributions are as follows:

 \begin{figure}[]
 \centering
 \includegraphics[scale=0.17]{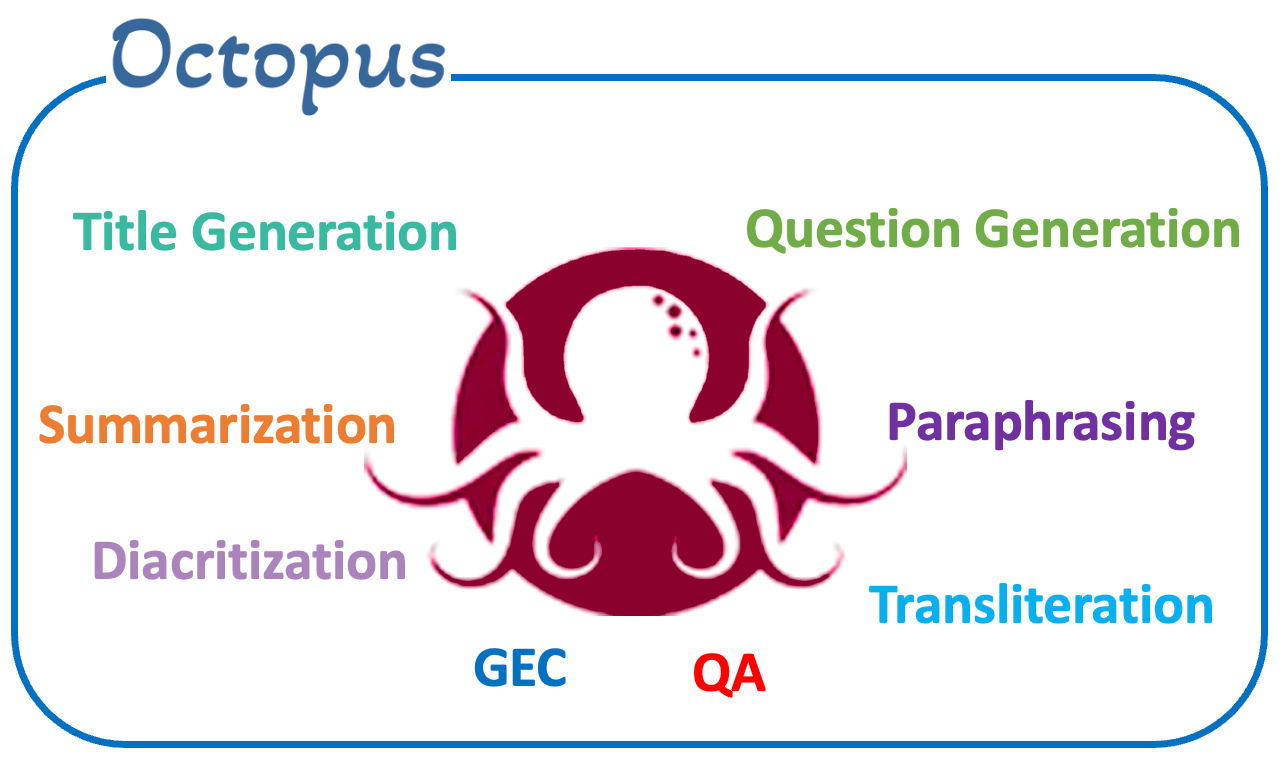}
  \caption{\octopus~is a jointly pretrained to cover eight NLG tasks, all shown in the illustration. } \label{fig:tasks} 
 \end{figure}

\begin{enumerate}

\item We pretrain better and faster-to-converge versions of the text-to-text transformer model AraT5, collectively dubbed \textit{AraT5\textsubscript{v2}}. Compared to~\newcite{nagoudi-etal-2022-arat5}, we train these new versions on a larger and more diverse dataset, as well as a larger sequence length.
\item To develop our models, we investigate diverse training strategies that integrate a combination of \textit{supervised} and \textit{unsupervised} training techniques.
    
\item  We introduce \octopus, a Python-based toolkit for eight Arabic NLG tasks. Our tool can be used as a strong baseline or as a core enabling technology that facilitates other developments. 
   
\item We will make \octopus~ publicly available to the research community.
\end{enumerate}


\section{Related Work}\label{sec:RW}

In the following section, we offer a concise overview of publicly available Arabic NLU and NLG tools, along with the Arabic and multilingual sequence-to-sequence (S2S) language models that we employ in this work.

\subsection{Arabic NLP Tools}
\noindent\textbf{NLU tools.}
Numerous attempts have been made to develop tools for assisting with Arabic. Some tools focus on aspects such as morphosyntax, encompassing tasks like morphological analysis, disambiguation, part-of-speech tagging, and diacritization. Notable examples include Stanford CoreNLP~\cite{manning2014stanford}, MADAMIRA~\cite{2014-madamira}, Farasa~\cite{2016-farasa}, and  CAMeL tools~\cite{camel2020}. Other tools, such as Mazajek \cite{farha2019mazajak}, and AraNet \cite{abdul2019aranet}, are dedicated to social meaning tasks such as sentiment analysis, emotion detection, age and gender prediction, and sarcasm detection.

\noindent\textbf{NLG Tools.} Regarding Arabic NLG, as far as we know, the only publicly available tools are primarily focused on many-to-Arabic machine translation (MT). These include OPEN-MT~\cite{EAMT2020}, NLLB~\cite{costa2022no}, and Turjuman~\cite{nagoudi-turjuman_2022}.

\subsection{Arabic S2S Language Model.}
 Here, we overview the Arabic sequence-to-sequence models we employ as baseline in this work.

\noindent\textbf{mT5.} This is the multilingual version  of T5 model~\cite{raffel2019exploring} introduced by~\newcite{xue2020mt5}. Pretraining of mT5 is performed on the extensive mC4 (Multilingual Colossal Clean Crawled Corpus) which covers $101$ languages, including Arabic. 

\noindent{\textbf{mT0.}} Developed by~\newcite{muennighoff2022crosslingual}, this is a group of S2S models ranging from $300$M to $13$B  parameters trained to investigate cross-lingual generalization through multitask fine-tuning. The models are finetuned from pre-existing mT5 ~\cite{xue2020mt5} multilingual language models using a cross-lingual task mixture called xP3. 

\noindent{\textbf{AraBART.}} Introduced by ~\cite{eddine2022arabart}, this is a pretrained encoder-decoder model designed specifically for abstractive summarization tasks in the Arabic language. AraBART follows the architecture of BART~\cite{lewis2019bart} and has been pretrained on a $73$GB  of Arabic text data. 

\noindent\textbf{AraT5.} Presented by~\newcite{nagoudi2022_arat5}, this is an  Arabic text-to-text Transformer model dedicated to MSA and Arabic dialects.  It is similar in configuration and size to T5~\cite{raffel2019exploring} and is trained on $248$GB of Arabic text ($70$GB MSA and $178$GB tweets). 
We now introduce our new model.

\section{AraT5\textsubscript{v2}}\label{sec:arat5}
\begin{table*}[!]
\centering
\resizebox{0.8\textwidth}{!}{%
\begin{tabular}{lccc|c}
\toprule
 & \textbf{AraT5\textsubscript{v1} }& \textbf{AraT5\textsubscript{v1}-MSA} & \textbf{AraT5\textsubscript{v1}-TWT} & \textbf{AraT5\textsubscript{v2}} \\ \midrule
Data size & $248$ GB & $70$ GB & $178$ GB & $250$ GB \\ 
Tokens count & $29$ B & $7.1$ B & $21.9$ B & $25.6$ B \\ 
Linguistic diversity & MSA, Tweets\textsuperscript{$\dagger$} & MSA & Tweets\textsuperscript{$\dagger$} & CA, DA, MSA \\ 
Sequence length & $512$ & $512$ & $512$ & $2,048$ \\ \bottomrule
\end{tabular}%
\caption{Comparison between AraT5\textsubscript{v1} and AraT5\textsubscript{v2} models. It is worth noting that our new model (AraT5\textsubscript{v2}) does not include tweets, whereas $71.77$\% of AraT5\textsubscript{v1} data is from Twitter (with the remaining $28.23$\% sourced from other sources). \textbf{CA:} Classical Arabic. \textbf{DA:} Dialectical Arabic. \textbf{MSA:} Modern Standard Arabic. Notably, Tweets\textsuperscript{$\dagger$} may encompass content in CA, DA, and MSA. }
\label{tab:comparison}
}

\end{table*}

In this section, we present a novel version of AraT5, the Arabic-specific sequence-to-sequence model. We refer to this novel version as AraT5\textsubscript{v2}. This new version represents a substantial evolution of the original AraT5\textsubscript{v1} model,\footnote{In this paper, we refer to the original AraT5~\cite{nagoudi-etal-2022-arat5} as  AraT5\textsubscript{v1}.} marked by notable improvements. These include \textbf{(1)}~training on an expanded dataset comprising both labeled and unlabeled data,  \textbf{(2)}~larger sequence length of $2,048$ tokens, and \textbf{(3)}~diverse training strategies that integrate a combination of unsupervised and supervised training techniques. Table~\ref{tab:comparison} provides a comparison between AraT5\textsubscript{v1} and AraT5\textsubscript{v2}. 

\noindent\textbf{Pretraining data.} As we mentioned previously, our pretraining (unlabeled and labeled) dataset is linguistically diverse, covering all categories of Arabic (i.e., CA, DA, and MSA). as we will now describe. 

\subsection{Unlabled Data} \label{sub_sec:lm_data} 

We collect approximately $250$GB of Arabic MSA text, which corresponds to around $25.6$B tokens.\footnote{We note that AraT5\textsubscript{v1} trained only on $70$GB MSA data.}  We use different sources including  AraNews\textsubscript{v2}~\cite{nagoudi2020machine}, El-Khair~\cite{elkhair-2016}, Gigaword,\footnote{\href{https://catalog.ldc.upenn.edu/LDC2009T30}{https://catalog.ldc.upenn.edu/LDC2009T30}.} OSIAN~\cite{zeroual2019osian}, Wikipedia Arabic,  Hindawi Books,\footnote{\href{https://www.hindawi.org/books/}{https://www.hindawi.org/books}.} OSCAR\textsubscript{Egyptian}~\cite{suarez2019asynchronous}, and AraC4~\cite{nagoudi2022jasmine}.\footnote{We note that AraC4 contains a diverse Arabic dialect as described in ~\cite{nagoudi2022jasmine}.} To obtain Classical Arabic (CA) data, we utilize the Open Islamicate Texts Initiative (OpenITI) corpus (v1.6)~\cite{nigst2020openiti}. The OpenITI corpus consists of $11$K  Islamic books, primarily collected from sources such as Shamela Library,\footnote{\href{https://shamela.ws}{https://shamela.ws}.} Al-Jami Al-Kabir collection (JK),\footnote{\href{http://kitab-project.org/docs/openITI}{http://kitab-project.org/docs/openITI}.} books digitized by the Jordanian publisher Markaz Al-Turāth, and the Shia Library.\footnote{\href{https://shiaonlinelibrary.com}{https://shiaonlinelibrary.com}.}

\subsection{Labeled Data} \label{sub_sec:labled_data} 
Recently, \newcite{nagoudi2023dolphin} introduced \textit{Dolphin}, an NLG benchmark for Arabic.  Dolphin covers MSA, Classical Arabic, and various Arabic dialects. It is composed of $40$ datasets, making it the largest and most diverse Arabic NLG benchmark. 
Due to the availability of the powerful Arabic machine translation toolkit, TURJUMAN~\cite{nagoudi-turjuman_2022}, we shift our focus away from machine translation, code-switching, and Arabization tasks in this paper. Hence, we utilize datasets from eight out of the total thirteen NLG tasks in Dolphin. In the following sections, we will provide a brief description of each of these tasks.

\noindent\textbf{{(1) Diacritization}.}  Is the computational procedure of adding missing diacritics or vowels to Arabic texts. For this task, we use the Arabic diacritization dataset presented by \newcite{fadel2019arabic}.

\noindent\textbf{{(2) Grammatical Error Correction}.}  The GEC task is centered around the analysis of written text with the aim of automatically identifying and correcting a range of grammatical errors. We use three GEC datasets: QALB 2014~\cite{mohit2014first},  QALB 2015~\cite{rozovskaya2015second}, and  ZAEBUC~\cite{habash-palfreyman-2022-zaebuc}.

\noindent\textbf{{(3) News Title Generation}.} 
The objective of this task is to generate a suitable headline for a given news article. To accomplish this, we use two datasets: Arabic NTG~\cite{nagoudi2022_arat5} and XLSum~\cite{hasan2021xlsum}.\footnote{We note that XLSum~\cite{hasan2021xlsum} contains news articles that are annotated with both summaries and titles. For the NTG task, we use the pairs of articles and titles used to create the training data.}

\noindent\textbf{{(4) Paraphrasing}.} 
In this task, we use four paraphrasing datasets: AraPara, a multi-domain Arabic paraphrase dataset~\cite{nagoudi2022_arat5}, ASEP, an Arabic SemEval paraphrasing dataset~\cite{cer2017semeval}, Arabic paraphrasing benchmark (APB)~\cite{alian2019towards}, and  TaPaCo \cite{scherrer-2020-tapaco}.\footnote{We use the Arabic part only of TaPaCo.}

\noindent\textbf{{(5) Question Answering}.} 
In this task, four publicly available extractive QA datasets are employed: ARCD~\cite{mozannar2019neural} and the Arabic part of the following three multilingual datasets: MLQA~\cite{lewis2019mlqa}, XQuAD~\cite{artetxe2020cross}, and 
TyDiQA~\cite{artetxe2020cross}.

\noindent\textbf{{(6) Question Generation}.} The goal of this task is to create simple questions that are pertinent to passages, along with their corresponding answers. For this, we utilize triplets consisting of \textit{passages}, \textit{answers}, and \textit{questions}, all extracted from the same QA datasets.

\noindent\textbf{{(7) Text Summarisation}.} This task includes five publicly available datasets, including both Arabic and multilingual data: MassiveSum~\cite{varab-schluter-2021-massivesumm}, XLSum~\newcite{hasan2021xlsum}, CrossSum~\cite{bhattacharjee2021crosssum}, ANT~\cite{chouigui2021arabic}, and 
 MarSum~\cite{inbook}.

\noindent\textbf{{(8) Transliteration}.}  This task involves converting words or text from one writing system to another while maintaining the original language's pronunciation and sound. Three datasets are used to create this component: ANETA~\cite{ameur2019anetac}, ATAR~\cite{Atar-2021}, and NETransliteration~\cite{merhav2018design}.

\section{Training Strategies}\label{sec:traini_Stra}
In this section, we describe the different strategies we use to pretrain and finetune AraT5\textsubscript{v2}.

\subsection{Unsupervised Pretraining.} \label{subsec:unsupervised}

Here, we focus on using only our unlabeled data (see Section~\ref{sub_sec:lm_data}) for pretraining our AraT5\textsubscript{v2}. The objective function does not rely on labels but instead imparts the model with transferable knowledge that can be effectively applied to various downstream tasks.  We follow \newcite{raffel2019exploring} in
using a masked language modeling ``\textit{span-corruption}'' objective.  This approach involves replacing consecutive spans of input tokens with a mask token, and the model is trained to reconstruct the masked tokens.

\subsection{Supervised Finetuning}  \label{sub_sec:super_vised_FT}

We use the labeled data (see Section~\ref{sub_sec:labled_data}) to finetune the AraT5\textsubscript{v2} models under two settings: (i)~\textit{single task} and (ii)~\textit{multitask} finetuning. 

\noindent\textbf{Single task finetuning.} We individually finetune our AraT5\textsubscript{v2} models on each of the eight NLG tasks we select from the Dolphin NLG benchmark~\cite{nagoudi2023dolphin}.  

\noindent\textbf{Multitask finetuning.} We additionally explore multitask learning~\cite{caruana1997multitask,ruder2017overview} using our AraT5\textsubscript{v2} models. This strategy involves training the model on several tasks concurrently, allowing the model and its parameters to be shared across all tasks. The ultimate goal is to enhance performance on each individual task over time. 
To indicate the intended task for the model, we incorporate a task-specific text ``\textit{prefix}'' to the original input sequence before it is fed into the model. For example, for the paraphrase task, the source will be: \u \textit{{paraphrase}:}\o~\small\textit{\<امرأة تضيف التوابل إلى اللحم>}. \normalsize \n The model should predict \g ~\small\textit{\<إمرأة تضيف المكونات إلى لحم البقر>}\normalsize \n.

\subsection{Joint Pretraining and Finetuning} \label{subsec:joint}
In this scenario, we establish a uniform training objective for both pretraining and finetuning. The model is trained using a maximum likelihood objective, employing ``\textit{teacher forcing}" \cite{raffel2019exploring,williams1989learning}, regardless of the specific task.

\section{Empirical Evaluation}\label{sec:eval}
\begin{table*}[!ht]
\centering
\resizebox{0.99\textwidth}{!}{%
\begin{tabular}{llcHccccHccHccc}
\toprule
\multirow{2}{*}{\textbf{Task}} & \multirow{2}{*}{\textbf{Test Set}} & \multirow{2}{*}{\textbf{Metric}} & \multirow{2}{*}{\textbf{}} & \multicolumn{4}{c}{\textbf{Baselines}} & \multirow{2}{*}{ \begin{tabular}[]{c}\bf AraT5\textsubscript{v2}\\ \bf1024\end{tabular}} & \multicolumn{2}{c}{\textbf{AraT5\textsubscript{v2}}} & \multirow{2}{*}{\textbf{}} & \multicolumn{3}{c}{\textbf{AraT5\textsubscript{v2}-Joint}} \\ \cmidrule{5-8} \cmidrule{10-11} \cmidrule{13-15}
 &  &  & & \textbf{mT0}& \textbf{mT5} & \textbf{AraBART} & \textbf{AraT5\textsubscript{v1}\textsuperscript{$\dagger$}} &  & \textbf{sTask} & \textbf{mTask\textsuperscript{$\star$}}  &  & \textbf{Joint\textsuperscript{$\star$}} & \textbf{sTask} & \textbf{mTask\textsuperscript{$\star$}} \\
\midrule

DIAC & ADT~\textbf{$\downarrow$} & \texttt{CER} & X & $1.58$\textsuperscript{±0.13} & $1.64$\textsuperscript{±0.11} & $23.43$\textsuperscript{±1.51} & $2.58$\textsuperscript{±0.19} & $1.55$\textsuperscript{±0.05} & \colorbox{green!20}{$\bf 1.30$\textsuperscript{±$0.20$}} & $1.97$ & E & $2.20$ & $1.90$\textsuperscript{±$0.24$} & $1.74$ \\ \midrule

\multirow{3}{*}{GEC} & QALB 2014 & \multirow{3}{*}{\texttt{F\textsubscript{0.5}} (\texttt{M\textsuperscript{2}})} & X & $65.86$\textsuperscript{±$0.67$} & $66.45$\textsuperscript{±$0.22$} & $68.67$\textsuperscript{±$0.08$} & $64.92$\textsuperscript{±$0.23$} & $70.57$\textsuperscript{±$0.19$} & $70.52$\textsuperscript{±$0.15$} & $62.36$ & E & $62.36$ & \colorbox{green!20}{$\bf 70.73$\textsuperscript{±$0.27$}} & $64.36$ \\ 
 & QALB 2015 L1 &  & X & $66.90$\textsuperscript{±$0.92$} & $66.68$\textsuperscript{±$0.08$} & $69.31$\textsuperscript{±$1.55$} & $64.22$\textsuperscript{±$0.82$} & \colorbox{green!20}{$\bf 71.28$\textsuperscript{±$0.18$}} & $70.8$\textsuperscript{±$0.12$} & $62.46$ & E & 62.46 & \colorbox{green!20}{$\bf 71.17$\textsuperscript{±$0.16$}} & $64.93$ \\ 
 & ZAEBUC & & X & $47.33$\textsuperscript{±$3.34$} & $46.90$\textsuperscript{±$0.87$} & $82.08$\textsuperscript{±$7.54$} & $75.78$\textsuperscript{±$2.43$} & $84.39$\textsuperscript{±$0.76$} & \colorbox{green!20}{$\bf 85.52$\textsuperscript{±$0.69$}} & $37.89$ & E & $42.25$ & $84.87$\textsuperscript{±$0.58$} & $78.30$ \\ \midrule
 
\multirow{3}{*}{PARA} & TAPACO & \multirow{3}{*}{\texttt{Belu}} & X & $15.43$\textsuperscript{±$0.64$} & $14.89$\textsuperscript{±$0.28$} & \colorbox{green!20}{$\bf 17.90$\textsuperscript{±$1.06$}} & $15.90$\textsuperscript{±$0.06$} & $16.49$\textsuperscript{±$0.34$} & $16.82$\textsuperscript{±$0.41$} & $11.73$ & E & $10.39$ & $18.14$\textsuperscript{±$0.84$} & $11.68$ \\ 
 & APB &  & X & \colorbox{green!20}{$\bf 38.36$\textsuperscript{±$0.14$}} & $24.29$\textsuperscript{±$13.98$} & $37.66$\textsuperscript{±$1.01$} & $20.34$\textsuperscript{±$1.82$} & $35.86$\textsuperscript{±$0.71$} & $35.04$\textsuperscript{±$0.89$} & $19.57$ & E & $16.92$ & $36.89$\textsuperscript{±$0.44$} & $16.93$ \\ 
 & SemEval &  & X & $20.49$\textsuperscript{±$0.13$} & $20.23$\textsuperscript{±$0.03$} & $24.52$\textsuperscript{±$0.62$} & $19.33$\textsuperscript{±$0.08$} & $24.59$\textsuperscript{±$0.25$} & $25.52$\textsuperscript{±$0.58$} & \colorbox{green!20}{$\bf 72.53$ }& E & $68.57$ & $27.02$\textsuperscript{±$0.53$} & $72.72$ \\ \midrule
 
\multirow{4}{*}{QA} & ARCD\textsubscript{QA} & \texttt{F\textsubscript{1}} & X & $53.24$\textsuperscript{±0.24} & $51.63$\textsuperscript{±$1.01$} & $50.26$\textsuperscript{±$0.99$} & $58.12$\textsuperscript{±$0.16$} & $61.53$\textsuperscript{±$0.24$} & $61.72$\textsuperscript{±$0.89$} & $55.43$ & E & $53.84$ & \colorbox{green!20}{$\bf 62.49$\textsuperscript{±$0.69$}} & $54.81$ \\ 
 & TyDiQA\textsubscript{QA} &  & X & $76.31$\textsuperscript{±$0.09$} & $74.99$\textsuperscript{±$0.23$} & $73.32$\textsuperscript{±$1.21$} & $39.55$\textsuperscript{±$1.96$} & $83.55$\textsuperscript{±$0.09$} & $82.99$\textsuperscript{±$0.47$} & $72.37$ & E & $71.72$ & \colorbox{green!20}{$\bf 84.21$\textsuperscript{±$0.47$}} & $72.44$ \\ 
 & XSQUAD\textsubscript{QA} &  & X & $54.55$\textsuperscript{±$0.76$} & $47.43$\textsuperscript{±$0.91$} & $47.33$\textsuperscript{±$0.87$} & $48.71$\textsuperscript{±$0.5$} & $59.16$\textsuperscript{±$0.46$} & $57.79$\textsuperscript{±$1.08$} & $63.73$ & E & $63.39$ & $59.42$\textsuperscript{±$0.72$} & \colorbox{green!20}{$\bf 64.89$} \\ 
 & LMQA\textsubscript{QA} &  & X & $49.17$\textsuperscript{±$0.34$} & $45.13$\textsuperscript{±$0.35$} & $47.24$\textsuperscript{±$0.13$} & $51.95$\textsuperscript{±$0.09$} & $54.88$\textsuperscript{±$0.18$} & $54.48$\textsuperscript{±$0.12$} & $47.50$ & E & $46.63$ & \colorbox{green!20}{$\bf 55.02$\textsuperscript{±$0.26$}} & $48.70$ \\ \midrule
 
\multirow{4}{*}{QG} & ARCD\textsubscript{QG} & \multirow{4}{*}{\texttt{Belu}} & X & $17.73$\textsuperscript{±$0.99$} & $17.62$\textsuperscript{±$2.1$} & $22.79$\textsuperscript{±$0.66$} & $16.8$\textsuperscript{±$1.32$} & $21.20$\textsuperscript{±$0.93$} & \colorbox{green!20}{$\bf 24.13$\textsuperscript{±$0.20$}} & $19.86$ & E & $19.23$ & $22.48$\textsuperscript{±$1.30$} & $21.54$ \\ 
 & TyDiQA\textsubscript{QG} &  & X & $30.22$\textsuperscript{±$0.91$} & $31.0$\textsuperscript{±$0.97$} & $33.64$\textsuperscript{±$0.13$} & $22.09$\textsuperscript{±$1.85$} & \colorbox{green!20}{$\bf 34.40$\textsuperscript{±$0.91$}} & $33.50$\textsuperscript{±$0.75$} & $25.37$ & E & $24.50$ & \colorbox{green!20}{$\bf 34.05$\textsuperscript{±$0.34$}} & $26.18$ \\ 
 & XSQUAD\textsubscript{QG} &  & X & $10.04$\textsuperscript{±$0.01$} & $9.96$\textsuperscript{±$0.03$} & $10.27$\textsuperscript{±$0.31$} & $9.21$\textsuperscript{±$0.09$} & \colorbox{green!20}{$\bf 11.71$\textsuperscript{±$0.32$}} & $10.98$\textsuperscript{±$6.91$} & $6.65$ & E & $1.94$ & \colorbox{green!20}{$\bf 11.50$\textsuperscript{±$0.41$}} & $7.30$ \\ 
 & MLQA\textsubscript{QG} &  & X & $6.04$\textsuperscript{±$0.08$} & $6.00$\textsuperscript{±$0.38$} & $7.02$\textsuperscript{±$0.09$} & $6.12$\textsuperscript{±$0.42$} & $7.32$\textsuperscript{±$0.14$} & \colorbox{green!20}{$\bf 7.56$\textsuperscript{±$0.27$}} & $3.96$ & E & $3.25$ & $7.28$\textsuperscript{±$0.11$} & $3.66$ \\ \midrule
 
\multirow{5}{*}{SUM} & XLSum & \multirow{5}{*}{ \texttt{Rouge\textsubscript{L}} } & X & $21.46$\textsuperscript{±$0.54$} & $20.64$\textsuperscript{±$0.31$} & $26.64$\textsuperscript{±$0.04$} & $22.71$\textsuperscript{±$1.36$} & $27.27$\textsuperscript{±$0.09$} & $27.15$\textsuperscript{±$0.09$} & $63.59$ & E & $52.25$ & $28.12$\textsuperscript{±$0.12$} & \colorbox{green!20}{$\bf 65.66$} \\ 
 & CrossSum & & X & $21.00$\textsuperscript{±$0.38$} & $20.29$\textsuperscript{±$0.01$} & $25.89$\textsuperscript{±$0.09$} & $22.14$\textsuperscript{±$1.53$} & $26.53$\textsuperscript{±$0.09$} & $26.57$\textsuperscript{±$0.06$} & $59.45$ & E & $50.82$ & $27.56$\textsuperscript{±$0.06$} & \colorbox{green!20}{$\bf 61.31$} \\ 
 & MarSum & & X & $23.00$\textsuperscript{±$0.17$} & $22.57$\textsuperscript{±$0.21$} & $26.49$\textsuperscript{±$0.03$} & $21.71$\textsuperscript{±$0.39$} & $26.28$\textsuperscript{±$0.06$} & $26.64$\textsuperscript{±$0.06$} & $20.49$ & E & $19.04$ & \colorbox{green!20}{$\bf 26.81$\textsuperscript{±$0.06$} }& $20.78$ \\ 
 & MassiveSum &  & X & $25.57$\textsuperscript{±$0.11$} & $22.88$\textsuperscript{±$0.12$} & $30.0$\textsuperscript{±$0.11$} & $15.89$\textsuperscript{±$0.4$} & $25.75$\textsuperscript{±$0.32$} & $23.00$\textsuperscript{±$0.00$} & $27.22$ & E & $25.75$ & \colorbox{green!20}{$\bf 27.69$\textsuperscript{±$0.07$}} & $26.97$ \\ 
 & ANTCorp &  & X & $90.29$\textsuperscript{±$0.11$} & $88.84$\textsuperscript{±$0.91$} & $90.0$\textsuperscript{±$0.20$} & $86.64$\textsuperscript{±$0.22$} & \colorbox{green!20}{$\bf 91.26$\textsuperscript{±$0.05$}} & $90.94$\textsuperscript{±$0.14$} & $87.39$ & E & $86.92$ & $90.85$\textsuperscript{±$0.12$} & $88.22$ \\ \midrule
 
\multirow{2}{*}{TG} & Arabic NTG & \multirow{2}{*}{\texttt{Bleu}} & X & $19.03$\textsuperscript{±$0.34$} & $19.23$\textsuperscript{±$0.01$} & $22.75$\textsuperscript{±$0.09$} & $19.55$\textsuperscript{±$0.16$} & $21.90$\textsuperscript{±$0.17$} & $22.13$\textsuperscript{±$0.08$} & $22.54$ &  & $21.33$ & $22.37$\textsuperscript{±$0.06$} & \colorbox{green!20}{$\bf 22.94$} \\ 
 & XLSum &  & X & $6.50$\textsuperscript{±$0.17$} & $6.51$\textsuperscript{±$0.11$} & $8.98$\textsuperscript{±$0.18$} & $7.44$\textsuperscript{±$0.11$} & $9.58$\textsuperscript{±$0.19$} & $9.59$\textsuperscript{±$0.17$} & $6.21$ &  & $5.91$ & \colorbox{green!20}{$\bf 9.82$\textsuperscript{±$0.14$} }& $6.11$ \\ \midrule

\multirow{3}{*}{TR} & ANTAEC~\textbf{$\downarrow$}& \texttt{CER} & X & $19.21$\textsuperscript{±$0.48$} & $18.93$\textsuperscript{±$0.30$} & $18.29$\textsuperscript{±$0.29$} & $20.74$\textsuperscript{±$0.17$} & \colorbox{green!20}{$\bf 17.19$\textsuperscript{±$0.02$}} & \colorbox{green!20}{$\bf 18.06$\textsuperscript{±$0.21$}} & $31.50$ & E & $33.00$ & $19.25$\textsuperscript{±$0.06$} & $31.66$ \\ 
 & ATAR ~\textbf{$\downarrow$}& \texttt{CER} & X & $16.79$\textsuperscript{±$0.15$} & $16.68$\textsuperscript{±$0.22$} & $17.70$\textsuperscript{±$0.05$} & $36.51$\textsuperscript{±$1.53$} & $15.35$\textsuperscript{±$0.43$} & $14.96$\textsuperscript{±$0.05$} & $33.63$ & E & $35.90$ & \colorbox{green!20}{$\bf 14.70$\textsuperscript{±$0.05$}} & $33.19$ \\ 
 & NETTrans & \texttt{Belu} & X & $55.70$\textsuperscript{±$0.18$} & $55.02$\textsuperscript{±$0.47$} & $54.15$\textsuperscript{±$0.75$} & $51.89$\textsuperscript{±$0.64$} & $58.26$\textsuperscript{±$0.33$} & \colorbox{green!20}{$\bf 58.33$\textsuperscript{±$0.70$}} & $43.69$ & E & $42.65$ & $57.81$\textsuperscript{±$0.66$} & $43.18$ \\ \midrule
 
 &  & \textbf{\texttt{H-Score}}~\textbf{$\uparrow$} & X & $37.01$ & $35.42$ & $39.86$ & $34.59$ & $41.99$ & $41.90$ & $41.41$ & E & $38.73$ & $42.56$ & \colorbox{orange!20}{$\bf 42.89$} \\ \cmidrule{3-15}
 &  & \textbf{\texttt{L-Score}}~\textbf{$\downarrow$} & X & $12.53$ & $12.42$ & $19.81$ & $19.94$ & $11.36$ & \colorbox{orange!20}{$\bf 11.44$} & $22.37$ & E & $23.70$ & $11.95$ & $22.20$ \\ \bottomrule

\end{tabular}%
}
\caption{Average of three runs of finetuned Arabic and multilingual models on~\octopus~test. \textbf{\texttt{L-Score}}: refers to the macro-average scores of tasks where a lower score $\downarrow$ is better. \textbf{\texttt{H-Score}}:  refers to the macro-average scores of tasks where a higher score $\uparrow$ is better. \octopus~task clusters taxonomy: (DIAC, Diacritization), (GEC, Grammatical Error Correction), (PARA, Paraphrase), (QA, Question Answering), (QG, Question Generation), (SUM, Summarization), (TG, News Title Generation), and (TR,  Transliteration). \textsuperscript{$\dagger$}We refer to vanilla AraT5~\cite{nagoudi-etal-2022-arat5} as AraT5\textsubscript{v1}. \textsuperscript{$\star$}For the \textit{joint} and \textit{multitask} models, we utilize the labeled data during the further pretraining phase. Consequently, we employ it only once, as opposed to the regular single fine-tuning, which involves three runs. \colorbox{green!20}{\textbf{Bold and green:}} best score in the individual task. \colorbox{orange!20}{\textbf{Bold and orange:}} best average scores over all tasks. }
\label{tab:test_results}

\end{table*}

\subsection{Baselines}\label{subsec:eval_baselines}
We evaluate our models across various scenarios, contrasting them with both multilingual and Arabic sequence-to-sequence pretrained language models. Specifically, we make use of mT5~\cite{xue2020mt5} and mT0~\cite{muennighoff2022crosslingual} as multilingual pretrained models; while comparing to AraBART~\cite{eddine2022arabart} and AraT5\textsubscript{v1}~\cite{nagoudi-etal-2022-arat5} as Arabic models. We evaluate our AraT5\textsubscript{v2} models (under different settings) and the selected baseline models on all eight NLG tasks (i.e., labeled data) described in Section~\ref{sub_sec:labled_data}.



\subsection{Experimental Setup}\label{subsec:eval_setup}


For our experiments, we have two settings: one for the pretrained models and another for models we finetuning. We now describe each of these settings.

\subsubsection{Pretrained Models} To pretrain our \textit{{AraT5\textsubscript{v2}}} model from scratch, we use the unsupervised pertaining strategy described in Section~\ref{subsec:unsupervised}. We pretrain for one million steps on a Google TPU POD \texttt{v3-128}.\footnote{\href{https://sites.research.google/trc/about/}{https://sites.research.google/trc/about/}} We employ a constant learning rate of \texttt{1e\textsuperscript{-3}} and a dropout rate of $0.1$. We use a batch size of $1,024$ with sequence length $2,048$. We further pretrain AraT5\textsubscript{v2} incorporating both unsupervised and supervised data (i.e., \textit{joint strategy}; see Section~\ref{subsec:joint}), with the same hyperparameters for an additional $200$K steps. We refer to the resulting model as \textit{{AraT5\textsubscript{v2}-joint}}.


\subsubsection{Single Task Finetuning}
We finetune both AraT5\textsubscript{v2} and AraT5\textsubscript{v2}-joint, as well as baseline models, on the eight NLG tasks ($20$ datasets) for $20$ epochs. We use a learning rate of \texttt{5e\textsuperscript{-5}}, a batch size of $8$, and a maximum sequence length of $512$.\footnote{For GEC, we use a maximum sequence length of $1,024$.} In all single task experiments, we consistently select the best checkpoint for each model based on performance on the respective development set. Subsequently, we report performance of each model on the respective test set.

\subsubsection{Multitask Finetuning}
We extend the pretraining of \textit{AraT5\textsubscript{v2}} and \textit{AraT5\textsubscript{v2}-joint} with labeled data by an additional $100$K steps for each model, all within the multitask finetuning setting. These experiments are conducted using a Google TPU POD \texttt{v3-128} with the same hyperparameters as the initial pretraining.\footnote{\textit{AraT5\textsubscript{v2}-mTask} trains for a total of $1.1$M steps, whereas  \textit{AraT5\textsubscript{v2}-joint-mTask} undergoes training for $1.3$M steps.} For model comparisons in the single task setting, we calculate the average of three runs of finetuned Arabic and multilingual models on the test sets of each task. However, for the joint and multitask models, we incorporate labeled data during the subsequent pretraining phase, employing a fixed number of steps—$200$K for the joint model and $100$K for the multitask model. As a result, we conduct a single evaluation run for these models due to the high computation costs.
\begin{table*}[!ht]
\centering
\resizebox{0.7\textwidth}{!}{%
\begin{tabular}{lr}
\toprule
\textbf{\textit{Input text}} &  \<الخيـل والليـل والبيـداء تعرفنـي *** والسيف والرمح والقرطاس والقلـم>  \\ \cmidrule{2-2}
\textbf{\textit{Target}} & \u \<الخَيْـلُ وَاللّيْـلُ وَالبَيْـداءُ تَعرِفُنـي *** وَالسّيفُ وَالرّمحُ والقرْطاسُ وَالقَلَـمُ> \\  \cmidrule{2-2} \n
\textbf{\textit{Multitask model}} &
\b \RL{وَالْقَلـمِ}
\r \RL{وَقِرْطَاسُ}
\g \RL{وَالرُّمْحُ}
\g \RL{وَالسَّيْفُ}
 *** 
\b \RL{تَعْرِفَنـي}
\r \RL{ وَالْبِيلادْ}
\b \RL{الْخَيـلِ وَاللَّيْل} 
\\ \cmidrule{2-2}
\textbf{\textit{Single task model}} &
\g \RL{وَالسّيْفُ وَالرُّمْحُ وَالْقِرْطَاسُ وَالْقَلَمُ}
***
\g \RL{الَخَيْلُ وَاللَّيْلُ وَالْبَيْدَاءُ تَعْرِفُنِي} 
\\ \midrule

\textbf{\textit{Input text}} & 
\begin{tabular}[c]{@{}r@{}}
\RL{إبراهيم بن كنيف النبهاني، شاعر إسلامي، اشتهر بأبيات له أولها } \\
\RL{تعز فإن الصبر بالحر أجمل *** وليس على ريب الزمان معول}\\
\RL{تناقلت كتب الأدب أبياته وهو من شعراء الحماسة.}
\end{tabular}
\\ \cmidrule{2-2}
\textbf{\textit{Target}} & 
\begin{tabular}[c]{@{}r@{}}
\u \RL{إبراهيمُ بن كُنَيْفٍ النَّبْهانِيُّ، شاعرٌ إسلاميٌّ، اشْتُهِرَ بِأَبْياتٍ لَهُ أَوَّلُها}\\
\u \RL{تَعَزَّ فَإِنَّ الصَّبرَ بالْحُرِّ أَجْمَلُ *** وَلَيْسَ عَلَى رَيْبِ الزَّمانِ مُعَوَّلُ}\\
\u \RL{تناقَلَتْ كُتُبُ الأدبِ أبياتَهُ وهُوَ مِنْ شُعَراءِ الحَماسَةِ.}
\end{tabular}
\\ \cmidrule{2-2}
\textbf{\textit{Multitask model}} &
\begin{tabular}[c]{@{}r@{}}

\g \RL{إسْلَامِيٌّ، اُشْتُهِرَ بِأَبْيَاتٍ لَهُ أَوَّلُهَا}
\r \RL{شَاعٌ}
\g \RL{ النَّبَهَانِيُّ،}
\r \RL{كَنِيسٍ}
\g \RL{إبراهيم بن}
\\
\g \RL{ وَلَيْسَ عَلَى رَيْبِ الزَّمَانِ مَعْوَلٌ}
 *** 
\g \RL{ أَجْمَلُ}
\r \RL{بِالْأَحْرِّ}
\g \RL{فَإِنَّ الصَّبْ}
\b \RL{تَعَزّ}

\\
\g \RL{الْحُمَاسَةِ.}
\r \RL{شُعَبَاءِ}
\g \RL{وَهُوَ مِنْ}
\b \RL{أَبْيَاتِهِ}
\g \RL{تَنَاقَلَتْ كُتُبُ الأَدَبِ }
\end{tabular} \\ \cmidrule{2-2}
\textbf{\textit{Single task model}} &
\begin{tabular}[c]{@{}r@{}}

\g \RL{إبْرَاهِيمُ بْنُ كُنَيْفٍ النَّبْهَانِيُّ، شَاعِرٌ إسْلَامِيٌّ، اُشْتُهِرَ بِأَبْيَاتٍ لَهُ أَوَّلُهَا}
\\
\g \RL{وَلَيْسَ عَلَى رَيْبِ الزَّمَانِ مُعَوَّلٌ}
***

\g \RL{ فَإِنَّ الصَّبْرَ بِالْحُرِّ أَجْمَلُ}
\b \RL{ تَعُزْ}
\\
\g \RL{تَنَاقَلَتْ كُتُبُ الأَدَبِ أَبْيَاتِهِ وَهُوَ مِنْ شُعَرَاءِ الْحِمَاسَةِ.}
 
\end{tabular}
\\ \bottomrule

\end{tabular}%
\caption{Examples of negative task interference in the \textbf{diacritization task}, both in a single-task and multitask.  \textbf{Color taxonomy}: ``\u \textbf{blue}\n'' refers to the original text, ``\r \textbf{red}\n'' denotes a word-level error, ``\b \textbf{light red}\n'' indicates a partial diacritization error on one more letter, and ``\g \textbf{green}\n'' signifies correctness. For single task, we use  ``\textit{AraT5\textsubscript{v2}-sTask}'' whereas we use ``\textit{AraT5\textsubscript{v2}-joint-mTask}'' model as the multitask model.}
\label{tab:diac_examples}
}
\end{table*}
\begin{table*}[!ht]
\centering
\resizebox{0.99\textwidth}{!}{%
\begin{tabular}{lr}
\toprule
\rowcolor{red!10!}  \multicolumn{2}{c}{\textbf{News Article}} \\ \midrule
\multicolumn{2}{c}{
\begin{tabular}[c]{@{}r@{}}

\rowcolor{orange!2!} \RL{أكد النجم البرازيلي نيمار مهاجم نادي الهلال أن الدوري السعودي بات أكثر قوة من الدوري الفرنسي مذكراً الجميع بتجربته في الأخيرعندما
} \\
\RL{انتقل إلى باريس سان جيرمان صيف 2017. وأوضح نيمار خلال مؤتمر صحفي مقام في بارا البرازيلية لدى سؤاله عن الدوري السعودي:
} \\
\RL{وأؤكد لك أن كرة القدم هي نفسها ، الكرة هي نفسها و يسجلون الأهداف و بالنظر إلى الأسماء فإن الدوري السعودي بات أقوى من الدوري 
} \\
\RL{الفرنسي. التدريبات هناك شديدة ونتعطش أنا وزملائي للفوز هناك بشكل كبير والتتويج مع الهلال. وأضاف: الجميع اعتقد أن الدوري السعودي
} \\
\RL{ضعيف والأمر نفسه حدث معي عندما انتقلت إلى الدوري الفرنسي، حينها ظن الناس الأمر نفسه لكني لم أضرب في حياتي من قبل المدافعين 
} \\
\RL{أكثرمن هناك. وأبان حول الدوري السعودي: اللاعبون الذين يلعبون هناك يعلمون مدى صعوبة اللعب في الدوري السعودي وأنا متأكد أنه لن
} \\
\RL{يكون أمرا سهلا الفوز بالمسابقة لأن الفرق عززت صفوفها بلاعبين جدد، وستكون بطولة ممتعة وشيقة جدا. وتلعب البرازيل أمام بوليفيا في
} \\
\RL{بارا البرازيلية يوم السبت قبل أن تواجه بيرو يوم الأربعاء ضمن تصفيات كأس العالم لمنتخبات أميركا الجنوبية. 
} \\

\end{tabular} 

}\\ \midrule

\rowcolor{red!10!}  \multicolumn{2}{c}{\textbf{Title Generation}} \\ \midrule
\g \textbf{Output} & \begin{tabular}[c]{@{}r@{}}
\g \RL{
نيمار: أعرف ماذا يعني اللعب في الدوري السعودي
}\\ 
\g \RL{
نيمار: الدوري السعودي أقوى من الفرنسي
}\\
\g \RL{
نيمار: أعرف ماذا يعني الفوز بالمباريات في الدوري السعودي
}\\
\g \RL{
نيمار: أعرف ماذا يعني أن الدوري السعودي أقوى من الفرنسي
}
\\
\g \RL{
نيمار: أعرف أن الدوري السعودي أقوى من الدوري الفرنسي
}
\end{tabular}  \\ \midrule

\rowcolor{red!10!}  \multicolumn{2}{c}{\textbf{Question Answering}} \\ \midrule 
\rowcolor{orange!2!}  \textbf{Question no. 1} & \RL{متي تقام مباراة بوليفيا و البرازيل؟} \\ 
\g \textbf{Output} &  \g \RL{السبت} \\ \midrule
\rowcolor{orange!2!}\textbf{Question no. 2}  &  \RL{ متي انتقل نيمار الي باريس سان جيرمان؟}  \\ 
\g \textbf{Output} & \g  \RL{صيف 2017} \\ \midrule

\rowcolor{red!10!}  \multicolumn{2}{c}{\textbf{Question Generation}} \\ \midrule
\rowcolor{orange!2!} \textbf{Answer} &  \RL{ تلعب البرازيل أمام بوليفيا في بارا البرازيلية يوم السبت} \\
\g \textbf{Output} & \g  \RL{ من يقابل البرازيل في تصفيات كأس العالم؟} \\ \bottomrule
\end{tabular}%
}
\caption{\octopus~output examples based on a randomly picked article from a news website. We prompt \octopus~to generate five potential titles, answers based on the questions, and questions for the provided answer.}
\label{tab:octpus_output_examples1}

\end{table*}
\subsection{Evaluation Metrics}\label{subsec:eval_metrics}

We present the results of our models and the baseline models independently on each task of evaluated datasets, using the relevant metric. We employ \texttt{Bleu}  score as an evaluation metric for paraphrase, question generation, title (i.e. headline news) generation, and sentence-level transliteration tasks. Additionally, we use \texttt{Rouge\textsubscript{L}}, \texttt{F\textsubscript{1}}, and \texttt{F\textsubscript{0.5}} (\texttt{M\textsuperscript{2}}) as evaluation metrics for summarization, question answering, and grammatical error correction, respectively. For diacritization and word-level transliteration datasets, we utilize the character error rate (\texttt{CER}) metric. We split the evaluation scores into ``\texttt{L-Score}'' where lower ↓ is better (e.g., \texttt{CER}) and ``\texttt{H-Score}'' where higher ↑ is better, i.e., \texttt{Bleu}, \texttt{F\textsubscript{1}}, \texttt{F\textsubscript{0.5}}, and \texttt{Rouge\textsubscript{L}}. 

\subsection{Results}\label{subsec:eval}

Table~\ref{tab:test_results} shows that our proposed models, across different settings, outperform the baseline models in $\sim90$\% of the individual test sets ($18$ out of $20$). Notably, AraT5\textsubscript{v2} significantly outperforms the vanilla AraT5\textsubscript{v1}~\cite{nagoudi-etal-2022-arat5} by  $7.3$ and $8.58$ points in terms of the macro-average scores for tasks where \textit{higher ($\uparrow$)} and  \textit{lower ($\downarrow$)} score is better, respectively. Furthermore, AraT5\textsubscript{v2} markedly outpaces the second-ranked baseline model, AraBART, by an average of $2.04$ ($\uparrow$) and $8.45$  ($\downarrow$) in the macro-average scores.


Additionally, the AraT5\textsubscript{v2}-joint single-task model achieves the highest score in $8$ out of $20$ ($\sim40$\%) for the individual tasks, followed by the AraT5\textsubscript{v2} models and the AraT5\textsubscript{v2}-joint multitask model, each achieving the best score in $4$ out of $20$ ($\sim20$\%) tasks. It is also noteworthy that AraBART and mT0 each obtain the best score in only one task.

\subsection{Discussion}\label{subsec:discuss}

\noindent  Exploring different pretraining settings allows us to derive unique insights. Examples of insights that can be gleaned from Table~\ref{tab:test_results} include:

\noindent\textbf{Addressing open-domain problems}. We observe that sequence-to-sequence models like T5 encounter challenges when tackling open-domain question-answering tasks. For example, the results on the MLQA dataset demonstrate notably low performance across all evaluated models.

\noindent\textbf{Handling lengthy sentences}. Multitasking proves effective in addressing challenges when working with long texts, such as paragraphs or documents. It significantly excels in tasks involving long sequences. For instance, paraphrasing text such as the SmEval dataset and abstractive summarization like ARCD and XLSum all include long sequences. Conversely, it does not lead to significant improvements in short-text paraphrasing, such as those at the sentence level in datasets like APPB and TAPACO.

\noindent\textbf{Negative task inference}. Notably, multitask training in our experiments has a negative impact on character-level tasks. For instance, we randomly select two examples from an Arabic poetry website\footnote{\href{https://poetry.dctabudhabi.ae/}{https://poetry.dctabudhabi.ae/}}, remove diacritics from the input text, and require both the AraT5\textsubscript{v2}-joint multitask and AraT5\textsubscript{v2} single task models to diacritize these examples. As shown in Table~\ref{tab:diac_examples}, the multitask model alters the words themselves, while the single task model preserves the input words (i.e., it focuses solely on adding diacritization to the character sequences).

\subsection{Performance Comparison}\label{subsec:eval_preformance}

One of our primary objectives in developing a new version of AraT5 is to improve the time required for the finetuning process (i.e., convergence time). Therefore, we conduct a comparison between AraT5\textsubscript{v1} and AraT5\textsubscript{v2}, as well as the baselines models in this respect. This allows us to analyze their computational efficiency and gain insights into their convergence behavior. To quantify this, we measure the required average time for convergence (in hours) and the average number of epochs needed to achieve convergence based on model results on development datasets. For a fair comparison, we finetune all models for a maximum of $20$ epochs across all tasks. Notably, the evaluation results carry on the average of three separate runs using three different seeds, thereby enhancing the robustness and reliability of our comparison. 

\noindent{\textbf{Convergence time.}} In general, we observe that AraBART and AraT5\textsubscript{v2} 
need on average  $12$ and $13$ epochs, respectively,   till convergence compared to AraT5\textsubscript{v1}, which needs an average of $16$ epochs to achieve the best performance. So, we notice that AraBART requires only $2.9$ hours to converge and achieve the optimal performance, while AraT5\textsubscript{v2}, and AraT5\textsubscript{v1}, need an average of  $3.77$  and $5.20$ hours, respectively, to reach the best score. So, we observe that AraT5\textsubscript{v2} is approximately $35.19\%$ faster than AraT5\textsubscript{v1} in terms of training times.\footnote{We note that AraBART requires a shorter time to converge due to the vocabulary size of the model (32K) compared to (110K) for AraT5-based models.}

\section{\octopus~Toolkit} \label{sec:octopus_tool}

\begin{table*}[!ht]
\centering
\resizebox{0.99\textwidth}{!}{%
\begin{tabular}{lr}
\toprule

\rowcolor{red!10!}  \multicolumn{2}{c}{\textbf{Paraphrase}} \\ \midrule
\textbf{Input} & \begin{tabular}[c]{@{}r@{}}
\rowcolor{orange!2!} \RL{ بالتزامن مع استنفار ليبيا لاسيما في الشرق جراء الإعصار "دانيال" الذي ضربها، وصلت العاصفة دانيال أيضا إلى مصر وضربت محافظتين في شمال البلاد،
} \\
 \RL{هما مرسى مطروح والإسكندرية فقد غطت العاصفة مناطق الحدود المصرية غرب البلاد، وشملت مدن السلوم وسيوة ومطروح والإسكندرية. وبينت
}\\
\RL{القنواتالمختلفة لصور الأقمار الصناعية شكل العاصفة التي طرقت الأبواب المصرية. وأظهرت طبقات من السحب المختلفة والأتربة المصاحبة لها.
} \\

\end{tabular}  \\\midrule
\g \textbf{Output} & \begin{tabular}[c]{@{}r@{}}

\g \RL{بالتزامن مع حالة التأهب في شرق ليبيا بسبب إعصار دانيال، وصلت العاصفة إلى مصر لتغطي محافظتين شمال البلاد، وهما مرسى مطروح والإسكندرية.
} \\
\g \RL{ وأظهرت مختلف صور الأقمار الصناعية شكل العاصفة التي ضربت ليبيا،  والتي غطت إلى حد كبير شرق مصر، ولا سيما في مدينتي سرت والسلوم،
} \\
\g \RL{كما أظهرت مختلف طبقات الغيوم والغابات المصاحبة لها.
}

\end{tabular}  \\ \midrule

\rowcolor{red!10!}  \multicolumn{2}{c}{\textbf{Summarization}} \\ \midrule

\textbf{Input} & \begin{tabular}[c]{@{}r@{}}
\rowcolor{orange!2!} \RL{اعترف نضال حسن، الذي يدافع عن نفسه، بقتل الجنود، متحججا بحماية المسلمين وعناصر طالبان في أفغانستان، ولكن القاضي العسكري رفض حجته
} \\

\RL{ 'بحماية الآخرين'. وإذا أدين حسن، البالغ من العمر 42 عاما، بقتل 13 شخصا وجرح آخرين فإنه سيواجه عقوبة الإعدام. ويعتبر الحادث الأكثر دموية
} \\
\RL{
من بين الهجمات غير القتالية التي وقعت في قاعدة عسكرية أمريكية.  وقال شهود عيان دخل في 5 نوفمبر/تشرين الثاني عام 2009 مصحة تعج بالجنود
}\\
\RL{
الذين كانوا ينتظرون أدوارهم إجراء فحوصات طبية أو التلقيح، ثم صعد على مكتب، وأطلق النار من سلاحين بيديه، دون توقف إلا لإعادة تعبئة السلاح.
} \\
\RL{
مواضيع قد تهمك نهاية وسيقدم ممثلو الادعاء أدلة تفيد بأن حسن مال إلى الأفكار المتطرفة، وكان يزور المواقع بحثا عن ّالجهاديين' وطالبان، ساعات قبل
}\\
\RL{
الهجوم. وكان الرائد حسن سيلتحق بالقوات الأمريكية في أفغانستان قبل أن ينفذ هجومه. 'عنف في مكان العمل' وصنفت وزارة الدفاع الأمريكية الحادث 
}\\
\RL{
باعتباره 'عنفا في مكان العمل' بدلا من تصنيفه 'عملا إرهابيا'، وهو ما أغضب عائلات الضحايا، حسب ما أفاد به مراسل بي بي سي، نك براينت، في 
}\\
\RL{
فروت هود. ويتوقع أن يدلي العديد من جرحى الحادث بشاهاداتهم أمام المحكمة. وسيواجه حسن عددا من ضحاياه في قاعة المحكمة لأنه سيتولى الدفاع
} \\
\RL{ عن
نفسه. وهو يستخدم كرسيا متحركا لأنه أصيب بالشلل، عندما أطلق عليه شرطي في القاعدة العسكرية النار.
}

\end{tabular}  \\ \midrule
\g \textbf{Output} &
\begin{tabular}[c]{@{}r@{}}
\g \RL{تنظر محكمة عسكرية أمريكية في وقت لاحق من اليوم في قضية الطبيب النفسي العسكري الأمريكي، نضال حسن، الذي اعترف بقتل 13 شخصا وإصابة} \\
\g \RL{
 أكثر من ثلاثين آخرين في إطلاق نار بقاعدة فورت هود منذ أربعة أعوام.
}
\end{tabular} 
\\ \midrule
\rowcolor{red!10!}  \multicolumn{2}{c}{\textbf{Grammatical Error Correction}} \\ \midrule
 \textbf{Input} &
\begin{tabular}[c]{@{}r@{}}
\rowcolor{orange!2!}\RL{لا زال كبير الشبيحه يظن ان ارواح وآلام الناس اقل كلفه من تخليه عن منصبه ، فلذلك اذا كان السوريون لا يرتضون بهذه المعادله المهينه ، فعليهم ان يهبوا
}\\
\RL{ هبه قويه واحده وياخذو حقوقهم من هذه العصابه عنوه ، اننا يا أحبائي ندفع ثمن اكثر من اربعين عام ومن الخنوع والذل والثمن سيكون غاليا 
}\\
\RL{ولكنه يستأهل هذه التضحيات        
}
\end{tabular} 
\\ \midrule
\g \textbf{Output} &
\begin{tabular}[c]{@{}r@{}}
\g \RL{لازال كبير الشبيحة يظن أن أرواح وآلام الناس أقل كلفة من تخليه عن منصبه ، فلذلك إذا كان السوريون لا يرتضون بهذه المعادلة المهينة ، فعليهم أن يهبوا} \\
\g \RL{هبة قوية واحدة ، ويأخذوا حقوقهم من هذه العصابة عنوة . إننا يا أحبائي ندفع ثمن أكثر من أربعين عام ، ومن الخنوع والذل ، والثمن سيكون غاليا ،
} \\
\g \RL{ولكنه يستأهل هذه التضحيات .
} \\
\end{tabular} 
\\ \midrule
 \bottomrule
\end{tabular}%
}
\caption{\octopus~output examples for \textit{grammatical error correction}, \textit{paraphrasing}, and \textit{summarization}.}
\label{tab:octpus_output_examples2}

\end{table*}
\subsection{Model Selection}
Our objective is to introduce a versatile language generation toolkit capable of handling a wide range of tasks, all within a single model. To achieve this goal, we have explored multiple training strategies, as described in Section~\ref{sec:traini_Stra}. Based on our empirical evaluations, we observe that finetuned  \textit{AraT5\textsubscript{v2}-joint} under the multitask setting outperforms the other models.\footnote{As Table~\ref{tab:test_results} shows, \textit{AraT5\textsubscript{v2}-joint-mTask} outperforms other models where a higher score is better and remains highly competitive in scenarios where a lower score is preferred.} Therefore, we utilize this model as the foundation for developing our~\octopus~toolkit (illustrated in Figure~\ref{fig:tasks}).

\subsection{Task Coverage} 
\octopus~is designed for \textit{eight} machine generation tasks, encompassing diacritization, grammatical error correction, news headlines generation, paraphrasing, question answering, question generation, and
transliteration. This comprehensive package includes a Python library along with associated command-line scripts. Table~\ref{tab:octpus_output_examples1} illustrates the output of \octopus, generating five potential titles, answers derived from questions related to the content, and questions corresponding to a provided answer based on a randomly selected article from a news website. Moreover, Table~\ref{tab:octpus_output_examples2} showcases examples of \octopus~for grammatical error correction, paraphrasing, and summarization. We now describe the intricacies of implementation and design of the \octopus~toolkit, along with its various configurable settings. 

\subsection{Implementation}\label{subsec:implementation}
\begin{table*}[!ht]
\centering
\resizebox{0.99\textwidth}{!}{%
\begin{tabular}{lll}
\toprule
                                          & \multicolumn{1}{l}{\textbf{Argument}} & \multicolumn{1}{l}{\textbf{Description}}                          \\ \midrule
\multirow{3}{*}{\textbf{Basic}}           & \textit{- - help} {[}\textit{-h}{]}                        & To display the arguments details                                   \\ 
                                          & \textit{- - cache-dir} {[}\textit{-c}{]}                  & Specify the path to the cache directory.                           \\ 
                                          & \textit{- - logging-file} {[}\textit{-l}{]}               & Define the file path for logging.                                  \\ \midrule
\textbf{Task} &
  \textit{- - prefix} {[}\textit{-p}{]} &
  \begin{tabular}[c]{@{}l@{}}Task prefix should be one of the following: {[}'\textit{diacritize}', '\textit{correct\_grammar}',\\ '\textit{paraphrase}',  '\textit{answer\_question}',  '\textit{generate\_question}', '\textit{summarize}','\textit{generate\_title}', \\ '\textit{translitrate\_ar2en}', '\textit{translitrate\_en2ar}' {]}\end{tabular} \\ \midrule
\multirow{5}{*}{\textbf{Input \& Output}} & \textit{- - text }{[}\textit{-t}{]}                        & Provide the input text for generative tasks.                       \\ 
                                          & \textit{- - input-file} {[}\textit{-f}{]}                 & Specify the path of the input file.                                \\ 
                                          & \textit{- - max-outputs} {[}\textit{-o}{]}                & Define the number of hypotheses to generate as output.             \\ 
                                          & \textit{- - batch-size} {[}-bs{]}                & Set the number of input sentences processed in a single iteration. \\ 
                                          & \textit{- - seq-length }{[}\textit{-s}{]}                 & Specify the maximum sequence length for the generative text.       \\ \midrule
\multirow{5}{*}{\textbf{Decoding}} &
  \textit{- - search-method} {[}\textit{-m}{]} &
  Choose the decoding method from the options {[}‘\textit{greedy}’, ‘\textit{beam}’, ‘\textit{sampling}’{]}. \\ 
                                          &\textit{ - - nbeam} {[}\textit{-nb}{]}                    & If using beam search, specify the beam search size.      \\ 
                                          & \textit{- - no-repeat-ngram-size} {[}\textit{-ng}{]}    & Avoid repeating the same n-gram size in the generated text.        \\ 
                                          & \textit{- - top-k} {[}\textit{-k}{]}                      & Utilize sampling with a top-k strategy.                            \\ 
                                          & \textit{- - top-p} {[}\textit{-p}{]}                      & Implement sampling with a top-p strategy.                          \\ \bottomrule
\end{tabular}%
}
\caption{\octopus~command line argument list.}
 \label{tab:agrs}
\end{table*}

We distribute \octopus~as a modular toolkit built using standard libraries including \texttt{PyTorch}~\cite{paszke2019pytorch} and \texttt{HuggingFace}~\cite{lhoest2021datasets}. It is implemented in Python and can be easily installed using the \texttt{pip} package. It is compatible with Python versions 3.8 and later, \texttt{Torch} version 2.0 and later, and the \texttt{HuggingFace Transformers} library version 4.30 or higher.\footnote{Installation instructions and documentation can be found at:~\href{https://github.com/UBC-NLP/octopus}{https://github.com/UBC-NLP/octopus}.} We offer three usage options with varieties of arguments: \textit{(i) Command-Line Interface (CLI)}, \textit{(ii) Python integration package}, and \textit{(iii) an interactive web interface}.

\noindent\textbf{CLI ommands.} We offer three command-line interfaces for task selection and output generation as follows:
\noindent First, the ``\textit{octopus\_interactive}'' command provides an interactive mode that allows users to actively engage with the system. With this command, users can efficiently select their desired task and input text and then apply the chosen task to generate output. For instance, if a user wants to diacritize several sentences, they can initiate the diacritization task and input the sentences one by one to undergo the diacritization process.
Second, the main command ``\textit{octopus}'' offers two options: users can either directly input the text or specify a file path, allowing flexibility in applying multiple tasks to a large amount of data points. Finally, the task-specific command ``\textit{octopus-taskname}'' offers seven task-specific commands, each corresponding to one of the supported tasks. For instance, there are ``\textit{octopus-diacritize}'' and ``\textit{octopus-paraphrase}'' commands. These task-specific commands follow the same usage pattern as the ``octopus'' command, but are designed for individual tasks.

\noindent\textbf{Python integration package} \octopus~is a Python library that offers numerous functions for seamless integration with various dataframe architectures, including \texttt{Pandas}, \texttt{PySpark}, \texttt{Dask}, and more. It takes as input the function to be integrated into user code and returns both generative text and processing logs.

\noindent\textbf{Interactive web interface.} We offer a dynamic interactive web interface that allows users to try \octopus~tasks. Furthermore, to facilitate adoption, we provide a \texttt{Google Colab} notebook with detailed instructions on how to use the \octopus~tool and model, and integrate them with user's code.

\subsection{Arguments}\label{subsec:arguments}

Each of the command lines (i.e., \textit{octopus-interactive, octopus, or octopus-taskname} supports or requires several arguments. Furthermore, \octopus~supports four decoding methods on the decoder side: \textit{greedy search}, \textit{beam search}~\cite{koehn2009statistical}, \textit{top-k sampling}~\cite{fan2018hierarchical}, and \textit{nucleus sampling}~\cite{holtzman2019curious}. We set as the default setting \textit{beam search} with a beam size of $5$, and a maximum sequence length of $2,048$. Table~\ref{tab:agrs} shows detailed descriptions of the arguments and their usage. This information helps users understand and utilize the provided arguments effectively. 

\section{Conclusion}\label{sec:conc}
We introduced a suite of powerful Arabic text-to-text Transformer models trained on large and diverse datasets, with an extended sequence length of up to $2,048$. We also explored various pretraining strategies, including unsupervised and joint pertaining,  using both single and multitask settings. Our models outperform competitive baselines, demonstrating their effectiveness. Furthermore, we introduced \octopus, a publicly available Python-based package and command-line toolkit tailored for \textit{eight} Arabic natural language generation tasks. \octopus~is designed to be extensible, and we plan to expand its capabilities by adding more tasks and increasing the capacity of our back-end model. 
\section{Limitations}\label{sec:limitations}
We identify the following limitations:
\begin{itemize}
    \item \textbf{Dialectal Arabic}. In this paper, our primary focus is on MSA tasks. Nevertheless, we are committed to expanding our scope to cover tasks in available Arabic dialects in the future. Currently, there is a recognized necessity within the community to facilitate the creation of datasets tailored to multiple Arabic dialects. For example, there is currently a deficiency in dialectal resources for sequence-to-sequence tasks such as summarization, paraphrasing, and question-answering. As more resources are created for dialects covering these tasks, we anticipate enhancing the coverage and capabilities of \octopus~exploiting these resources. Fortunately, our toolkit and core back-end models are extensible and hence would allow for such a development seamlessly.

    \item \textbf{Task Coverage}. \octopus~currently encompasses only eight generation tasks. However, we have plans to expand its capabilities by including additional tasks. These upcoming additions can involve, for example, dialgoue geeration and tasks involving code-switching. Again, adding more tasks to~\octopus~will not be onerous, once respective datasets are available. 
    \item \textbf{Intended Use}.~\octopus~is a natural language generation toolkit designed to handle eight different tasks. We have tried the toolkit under different scenarios and found it to perform well. However, before any real-world usecases, we strongly encourage further and more extensive evaluations under diverse conditions.

\end{itemize}

\section{Ethical Considerations}\label{sec:ethics}

Our pretraining datasets are sourced from the public domain. Similarly, the labeled datasets used for model finetuning have been collected from publicly available data, made possible through the dedicated efforts of numerous researchers over the years. Consequently, we do not have significant concerns regarding the retrieval of personal information from our trained models. It is essential to note that the datasets we gather to construct \octopus~may contain potentially harmful content. Furthermore, during model evaluation, there is a possibility of exposure to biases that could lead to unintended content generation. For release, all our pretrained models and the toolkit are publicly available for non-malicious use.

\section*{Acknowledgments}\label{sec:acknow}
We acknowledge support from Canada Research Chairs (CRC), the Natural Sciences and Engineering Research Council of Canada (NSERC; RGPIN-2018-04267), the Social Sciences and Humanities Research Council of Canada (SSHRC; 435-2018-0576; 895-2020-1004; 895-2021-1008), Canadian Foundation for Innovation (CFI; 37771), Digital Research Alliance of Canada,\footnote{\href{https://alliancecan.ca}{https://alliancecan.ca}} and UBC ARC-Sockeye.\footnote{\href{https://arc.ubc.ca/ubc-arc-sockeye}{https://arc.ubc.ca/ubc-arc-sockeye}} We thank the
Google TFRC program for providing us with free TPU access.\footnote{\href{https://sites.research.google/trc/about/}{https://sites.research.google/trc/about/}}


\normalem
\bibliography{ubc-dlnlp}
\bibliographystyle{acl_natbib}

\end{document}